\documentclass{article} 
\usepackage{iclr2027_conference,times}

\usepackage{amsmath,amsfonts,bm}

\def\eqref#1{equation~\ref{#1}}

\def\1{\bm{1}}

\DeclareMathAlphabet{\mathsfit}{\encodingdefault}{\sfdefault}{m}{sl}
\SetMathAlphabet{\mathsfit}{bold}{\encodingdefault}{\sfdefault}{bx}{n}

\usepackage{hyperref}
\usepackage{url}
\usepackage{caption}

\usepackage{amsmath}
\usepackage{mathtools}
\usepackage{amsthm}
\usepackage{multirow}
\usepackage{adjustbox}
\usepackage{xcolor}
\usepackage{colortbl}

\usepackage[utf8]{inputenc} 
\usepackage[T1]{fontenc}    
\usepackage{hyperref}       
\usepackage{url}            
\usepackage{booktabs}       
\usepackage{amsfonts}       
\usepackage{nicefrac}       
\usepackage{microtype}      
\usepackage{xcolor}         
\usepackage{bm}
\usepackage{listings}

\definecolor{darkblue}{RGB}{44, 62, 80}      
\definecolor{mediumblue}{RGB}{52, 152, 219}   
\definecolor{lightblue}{RGB}{174, 214, 241}   
\definecolor{paleblue}{RGB}{235, 245, 251}    
\definecolor{grayblue}{RGB}{149, 165, 166}    

\definecolor{audioblue}{RGB}{230, 240, 255} 
\definecolor{visualgray}{RGB}{240, 240, 240} 

\definecolor{darkblue}{RGB}{44, 62, 80}      
\definecolor{mediumblue}{RGB}{52, 152, 219}   
\definecolor{lightblue}{RGB}{174, 214, 241}   
\definecolor{paleblue}{RGB}{235, 245, 251}    
\definecolor{grayblue}{RGB}{149, 165, 166}    

\definecolor{audioblue}{RGB}{230, 240, 255} 
\definecolor{visualgray}{RGB}{240, 240, 240} 

\title{Chain of Modality: Understanding and Controlling Modality Organization Bias in Omni-MLLMs }

\author{Ziyang Luo, Nian Liu, Junwei Han\\
Northwestern Polytechnical University\\
}

\iclrfinalcopy 
\begin{document}

\maketitle

\begin{abstract}
Omni-modal Large Language Models (Omni-MLLMs) are designed to reason over diverse sensory streams within a unified model. However, we find that multimodal reasoning is not solely determined by sensory evidence, but also by how modality information is organized. Through systematic analysis across diverse reasoning scenarios, we show that different modality organizations exhibit distinct advantages and limitations, and no single topology is universally optimal. Rather than proposing a general-purpose improvement to Omni-MLLM accuracy, we ask a narrower question: can these topology-induced failures be systematically identified and corrected? Motivated by this, we propose Chain of Modality (CoM), a framework that dynamically reorganizes multimodal topology during inference and learns adaptive organization strategies. Experiments across five benchmarks, diverse architectures, and model scales show that CoM reliably recovers topology-sensitive failures, through both training-free planning and lightweight Planner-SFT, while preserving performance on the full benchmark.

\end{abstract}


\vspace{-0.2cm}
\section{Introduction}
\vspace{-0.2cm}
The emergence of Omni-modal Large Language Models (Omni-MLLMs) has established a promising paradigm for unified perception and reasoning across heterogeneous modalities, including audio, vision, and text \cite{xu2025qwen2, li2025baichuan, xu2025qwen3, ai2025ming, team2026qwen3}. By jointly modeling multiple sensory streams within a unified framework, recent Omni-MLLMs have demonstrated increasingly strong multimodal perception and reasoning capabilities. A substantial body of recent work has therefore focused on improving the ability of Omni-MLLMs to understand and reason over heterogeneous modalities. 
\begin{figure}[!t]
    \centering
    \includegraphics[width=1\linewidth]{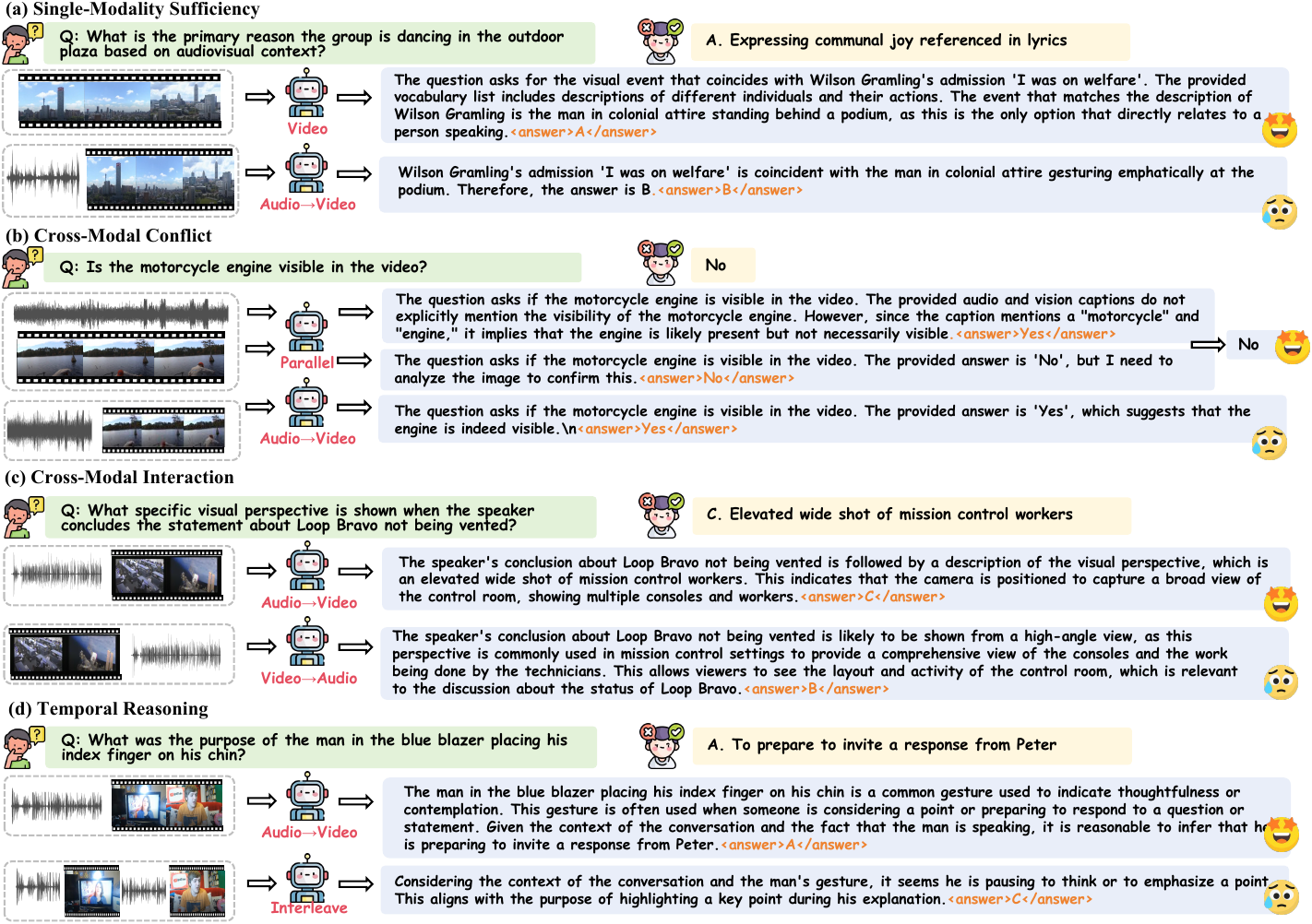}
    \vspace{-0.6cm}
    \caption{\textbf{Modality organization affects Omni-MLLM reasoning.}
Across diverse reasoning scenarios, different topology structures exhibit distinct advantages, motivating adaptive modality organization during inference.}
    \label{fig:motivation}
    \vspace{-0.85cm}
\end{figure}

Existing efforts mainly enhance multimodal reasoning through large-scale multimodal training, modality alignment, and post-training optimization. Specialized models enhance modality interaction through task-specific adaptation \cite{du2025crab, kim2025question, ye2024cat}, and generalist Omni-MLLMs improve multimodal reasoning through large-scale training and architectural scaling \cite{guan2026thinkomni, tao2025omnizip, li2026omnigaia}. Alignment-based approaches using reinforcement learning further optimize model behavior with preference- or task-level supervision \cite{chen2025omnidpo, ye2025eyes, kulkarni2025avatar, lu2025av, chen2026omnivideo}, and emerging agent frameworks improve complex task solving by orchestrating external tools such as ASR, captioning, and retrieval \cite{tao2025omniagent, li2026omnigaia, zhou2025daily}. More recently, \cite{selvakumar2026audio} questioned whether audio-visual LLMs truly exploit their available modalities, revealing modality bias and incomplete modality utilization. 
However, these studies primarily analyze \textit{which modalities} models rely on, while the role of \textit{how modalities are organized} during inference has received substantially less attention. In particular, existing Omni-MLLMs typically adopt a predefined multimodal input organization, leaving modality ordering and structural arrangement outside the inference decision process.

In this work, we revisit multimodal inference from the perspective of modality organization as shown in Figure~\ref{fig:motivation}. 
Prior studies have examined modality bias and input-order sensitivity in multimodal models \cite{tan2024order, zheng2025mllms, schilcher2025characterizing}, but this evidence comes largely \textit{from image-text settings only with sequential input}. We instead investigate whether the organization of audio-visual information affects multimodal reasoning in Omni-MLLMs.
To systematically study this question, we consider several representative multimodal reasoning scenarios, including single-modality sufficiency, cross-modal conflict, cross-modal interaction, and temporal reasoning. 
We examine each requirement under different modality organizations in Section~\ref{sec:prove}, including Single, Parallel, Sequential, and Interleaved organizations. Our experiments reveal that the benefit of multimodal information depends on how it is organized: different reasoning requirements exhibit different preferences over modality topologies, and no single organization consistently performs best across cases. These observations suggest that modality organization is not merely an input-format choice, but an important factor in Omni-modal inference.

To investigate whether and to what extent this phenomenon can be controlled, we develop a topology-controlled inference framework, termed \textbf{C}hain \textbf{o}f \textbf{M}odality (CoM). CoM consists of a Planner and a Reasoner, with an optional Decider. The Planner constructs modality chains by determining modality combinations and their interaction topologies, including Single, Parallel, Sequential, and Interleaved structures. 
The Reasoner performs inference following the specified organization. When multiple parallel chains are generated, the Decider aggregates their predictions through logit-level fusion to obtain the final prediction.

To evaluate the role of modality organization, we construct topology-sensitive subsets from fine-grained audio-visual perception benchmarks \cite{li2022learning}, open-domain multimodal scenarios \cite{zhou2025daily, li2024omnibench, hong2025worldsense}, and cross-modal hallucination evaluation \cite{sung2024avhbench}. 
These subsets capture two types of topology-sensitive cases: 
(1) modality integration failures, where unimodal organization produces the correct answer, while all multimodal organizations fail;
and (2) topology-induced prediction variance, where identical multimodal evidence produces different outcomes under different organizations.
On these challenging samples, we investigate two adaptation settings: training-free topology reorganization and planner-only supervision.
Training-free CoM shows that explicitly modifying modality organization can already improve multimodal inference, revealing the importance of topology selection itself. Further Planner-SFT  supervision achieves additional improvements and generalizes across unseen benchmarks, suggesting that the planner learns transferable organization preferences rather than dataset-specific heuristics. \textbf{Rather than focusing primarily on improving general Omni-MLLM accuracy, we focus on understanding and addressing topology-induced failures by establishing that they are systematic, identifiable, and learnable.}

Our contributions are summarized as follows:
\textbf{(1)} We reveal modality organization as an overlooked factor affecting Omni-MLLM reasoning, and show that different multimodal reasoning scenarios exhibit distinct preferences over modality topologies.
\textbf{(2)} We perform controlled intervention experiments to analyze the structural factors underlying topology sensitivity and introduce Chain of Modality (CoM) as a controllable framework for explicitly modeling and studying modality organization in multimodal inference.
\textbf{(3)} 
We investigate the controllability of modality organization through training-free topology reorganization and Planner-SFT, demonstrating that CoM can effectively learn and leverage adaptive organization strategies.

\vspace{-0.2cm}
\section{Related Work}
\vspace{-0.2cm}
\paragraph{Omni-modal Large Language Models.}
Omni-modal Large Language Models (Omni-MLLMs) integrate audio, vision, and text for general-purpose reasoning, spanning early domain-specific benchmarks \cite{yang2022avqa, li2022learning} to open-domain, hallucination-aware evaluations \cite{gong2024av, hong2025worldsense, zhou2025daily, sung2024avhbench}, and progressing from proprietary systems \cite{hurst2024gpt, comanici2025gemini} to open-source models such as Qwen-Omni \cite{xu2025qwen2, xu2025qwen3}, Baichuan-Omni \cite{li2025baichuan}, and Ola \cite{liu2025ola}. In practice, these models organize audio and visual streams either sequentially \cite{ai2025ming, liu2025ola} or as interleaved, timestamp-synchronized chunks \cite{xu2025qwen2, xu2025qwen3}, with the topology fixed in advance rather than adapted to the task.
Separately, prior work has shown that multimodal models are sensitive to input order \cite{tan2024order, zheng2025mllms, schilcher2025characterizing} and that multimodal fusion can produce cross-modal hallucination \cite{sung2024avhbench, leng2024curse}, with concurrent efforts auditing order sensitivity more broadly \cite{paruchuri2026same, xu2026systematic}. However, these studies mainly analyze the symptoms of modality bias or hallucination from image-text or multi-video settings, while the role of modality organization and interleaving topology remains largely unexplored. To our knowledge, no prior work studies whether these topology-dependent behaviors emerge in audio-visual Omni-MLLMs, where sequential or interleaved organizations are already adopted as default input paradigms.

\vspace{-0.2cm}
\paragraph{Omni-modal Reasoning and Adaptation.}
Beyond developing Omni-MLLM architectures, recent studies have explored various strategies to improve the reasoning capability of omni-modal systems. Specialized models, such as Crab \cite{du2025crab}, QA-TIGER \cite{kim2025question}, and CAT \cite{ye2024cat, ye2025cat+}, improve modality-specific understanding through task-oriented adaptation. Large-scale reasoning-oriented models, such as ThinkOmni \cite{guan2026thinkomni}, improve omni-modal reasoning through advanced inference strategies. Alignment-based approaches using DPO or reinforcement learning \cite{chen2025omnidpo, ye2025eyes, kulkarni2025avatar, lu2025av, chen2026omnivideo} further optimize model behaviors through preference or task-level supervision. Meanwhile, omni-modal agent frameworks, such as OmniAgent \cite{tao2025omniagent}, and OmniGaia \cite{li2026omnigaia}, improve complex task solving by orchestrating external tools or specialized modules. 
However, these analyses mainly focus on whether modalities are effectively perceived and utilized, treating modality organization as a fixed preprocessing choice rather than a controllable reasoning factor.
In contrast, our work investigates modality organization as an independent factor affecting Omni-MLLM reasoning and introduces the Chain of Modality (CoM) framework to explicitly model and study modality interactions.

\vspace{-0.2cm}
\section{Understanding Topology Preferences across Multimodal Reasoning Scenarios}
\vspace{-0.2cm}
\label{sec:prove}
To systematically study how modality organization interacts with different reasoning scenarios, we construct four controlled subsets: \textit{single-modality sufficiency}, \textit{cross-modal conflict}, \textit{cross-modal interaction}, and \textit{temporal reasoning}. Here, \textit{single-modality sufficiency} refers to cases where the answer can be correctly determined from either audio or visual information alone. \textit{Cross-modal conflict} refers to cases where the audio and visual streams describe different or inconsistent content. Moreover, \textit{cross-modal interaction} refers to cases where the two modalities provide complementary evidence, and correctly answering the question requires jointly integrating information from both. Finally, \textit{temporal reasoning} involves complementary multimodal evidence that additionally requires identifying or associating the relevant information with specific moments in time.
Rather than following benchmark-level categories, we categorize samples according to the type of evidence required for correct reasoning. Specifically, we draw candidate samples from representative benchmarks that contain potential cases of these reasoning patterns: Music-AVQA \cite{yang2022avqa}, AVHBench \cite{sung2024avhbench}, OmniBench \cite{li2024omnibench}, and DailyOmni \cite{zhou2025daily}.
We first use Qwen-Omni-30B \cite{xu2025qwen3} to pre-annotate candidate samples according to the evidence required for answering each question, based on the question, ground-truth answer, and multimodal content, without exposing predictions under different modality organizations. The annotations are then manually verified using predefined criteria, and ambiguous cases are removed. For each reasoning requirement, we retain about 200 verified samples to ensure balanced evaluation across different modality organizations.

\begin{table*}[t!]
\centering
\footnotesize
\setlength{\tabcolsep}{7pt}
\vspace{-0.2cm}
\caption{\textbf{Different multimodal reasoning scenarios exhibit distinct preferences for modality organization.} We fix the temperature and random seed across all experiments to reduce the impact of stochastic variation. ``-'' indicates unavailable results: Ola-7B does not support interleaved modality organization, and the cross-modal interaction subset only involves audio-image interaction.}
\vspace{-0.3cm}
\begin{tabular}{cc|cccccc}
\toprule
\multirow{2}{*}{\textbf{Dataset}} &\multirow{2}{*}{\textbf{Subset}}
 & \multicolumn{2}{c}{\textbf{Single}}& \multirow{2}{*}{\textbf{Parallel}} & \multicolumn{2}{c}{\textbf{Sequential}} & \multirow{2}{*}{\textbf{Interleaved}}\\
 & &\textbf{A} & \textbf{V} & & \textbf{A$\rightarrow$V} & \textbf{V$\rightarrow$A}\\
            \midrule
\rowcolor{paleblue} \multicolumn{8}{c}{\textit{\textbf{Qwen-Omni-7B}}}\\
\multirow{2}{*}{\textbf{single-modality sufficiency}} &Audio &\textbf{60.3} &27.6 &27.6 &39.7 &49.3 &55.2\\
&Video &35.7 &\textbf{88.6} &38.6 &82.9 &81.4 &80.0\\
\textbf{cross-modal conflict} &- &57.0 &67.0 &\textbf{68.0} &59.5 &61.0 &52.5\\
\multirow{1}{*}{\textbf{cross-modal interaction}} &- &31.3 &31.5 &34.8 &41.4 &\textbf{47.5} &-\\
\textbf{temporal reasoning} &- &35.6 &42.4 &47.2 &45.4 &47.7 &\textbf{48.5}\\
\rowcolor{paleblue} \multicolumn{8}{c}{\textit{\textbf{Ola-7B}}}\\
\multirow{2}{*}{\textbf{single-modality sufficiency}} &Audio &\textbf{57.4} &55.2 &56.6 &53.9 &56.6 &-\\
&Video &65.6 &\textbf{85.3} &80.3 &83.6 &77.1 &-\\
\textbf{cross-modal conflict} &- &67.2 &68.3 &\textbf{68.9} &56.7 &60.0 &-\\
\multirow{1}{*}{\textbf{cross-modal interaction}} &- &27.0 &30.0 &27.5 &35.0 &\textbf{37.0} &-\\
\textbf{temporal reasoning} &- &\textbf{44.3} &35.8 &39.8 &43.2 &41.5 &-\\
\bottomrule
\end{tabular}
\vspace{-0.3cm}
\label{tab:topology_sensitivity}
\end{table*}

We evaluate each controlled subset under four modality organizations: \textit{Single}, where only one modality is provided to inference; \textit{Parallel}, where audio and visual modalities are reasoned over independently and their predictions are subsequently aggregated; \textit{Sequential}, where modalities are concatenated sequentially following a specific order, such as Audio$\rightarrow$Visual or Visual$\rightarrow$Audio; and \textit{Interleaved}, where audio and visual segments are alternated to preserve fine-grained cross-modal correspondence. Table~\ref{tab:topology_sensitivity} summarizes the performance across the four reasoning scenarios. We observe that no single organization consistently dominates across all cases. Instead, different reasoning requirements exhibit distinct preferences over modality organization, suggesting that multimodal reasoning is determined not only by the available modalities but also by how they are structurally organized during inference.

\vspace{-0.3cm}
\paragraph{Single-modality Sufficiency}
As shown in Table~\ref{tab:topology_sensitivity}, we compare single-modality and multimodal organizations on both video-sufficient and audio-sufficient samples. For each subset, the modality containing decisive evidence achieves the best single-modality performance, while introducing the non-essential modality does not provide additional benefits and may even degrade performance. 
We further analyze modality-question interactions before and after introducing the non-essential modality in Table~\ref{tab:attention_shift}. Across both audio- and video-sufficient samples, introducing the non-essential modality shifts attention toward irrelevant evidence and reduces attention allocated to the decisive modality. These results suggest that unnecessary modalities may interfere with effective evidence utilization by altering the model's attention allocation across modalities.

\begin{table*}[t!]
\centering
\begin{minipage}{0.48\linewidth}
\centering
\caption{\textbf{Attention redistribution under single-modality sufficiency. }
Attention values denote question-to-modality attention.}
\vspace{-0.3cm}
\label{tab:attention_shift}
\resizebox{\linewidth}{!}{
\begin{tabular}{c|c|c|c}
\toprule
\textbf{Subset} 
& \textbf{Single} 
& \multicolumn{2}{c}{\textbf{A+V}} \\
\cmidrule(lr){3-4}
Qwen-Omni-7B& Decisive Attn
& Attn(A)
& Attn(V) \\
\midrule
\textbf{A-sufficient}
& Attn(A)=0.0487
& 0.0386
& 0.0103 \\
\textbf{V-sufficient}
& Attn(V)=0.0367
& 0.0489
& 0.0181 \\
\bottomrule
\end{tabular}
}
\end{minipage}
\hfill
\begin{minipage}{0.48\linewidth}
\centering
\caption{\textbf{Correct-answer average log-probability under cross-modal conflict.}
We compare unimodal and fused log-probability under different organizations.}
\vspace{-0.3cm}
\label{tab:conflict_logits}
\resizebox{\linewidth}{!}{
\begin{tabular}{c|c|c|c|c}
\toprule
\textbf{Subset}
&\textbf{Single-A}
& \textbf{Single-V}
& \textbf{Parallel}
& \textbf{Joint}\\
\midrule
\textbf{Conflict}
& -1.05
& -0.74
& \textbf{-0.67}
& -0.86\\
\bottomrule
\end{tabular}
}
\end{minipage}
\vspace{-0.4cm}
\end{table*}

\vspace{-0.3cm}
\paragraph{Cross-modal Conflict}
When audio and visual modalities provide conflicting evidence, Parallel organization achieves more reliable performance, as shown in Table~\ref{tab:topology_sensitivity}. The conflict subset contains samples with different modality biases, and neither unimodal branch consistently dominates across all samples. 
Unlike single-modality inference, conflict cases require integrating competing evidence from multiple modalities rather than relying on a single sensory stream.
We hypothesize that Parallel improves robustness by preserving modality-specific reasoning before evidence aggregation, allowing each modality to independently contribute its evidence.
To analyze this effect, we measure the average log-probability assigned to the ground-truth answer tokens under each organization (Table~\ref{tab:conflict_logits}). 
The two unimodal branches provide different confidence signals, and Parallel aggregation further improves the confidence of correct predictions. 
In contrast, joint organizations exhibit larger deviations after early cross-modal interaction, suggesting that premature fusion may cause conflicting evidence to interfere before modality-specific evidence is properly integrated.

\begin{figure}[!t]
    \centering
    \includegraphics[width=1\linewidth]{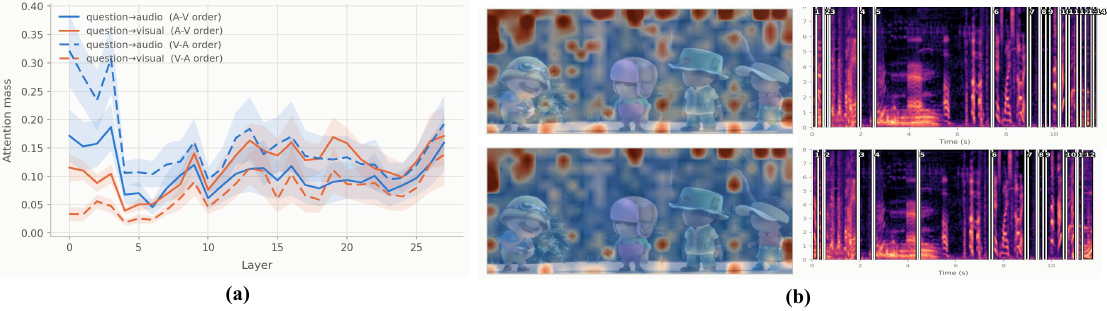}
    \vspace{-0.5cm}
    \caption{\textbf{(a) Layer-wise question-to-audio/video attention under different modality orders. (b) Attention visualization before and after modality reordering.} For audio inputs, white regions denote attended temporal regions. Up: $A \to V$, Down: $V \to A$}
    \label{fig:attention_change}
    \vspace{-0.4cm}
\end{figure}

\vspace{-0.3cm}
\paragraph{Cross-modal Interaction}
When audio and visual modalities provide complementary evidence, we observe that Sequential organization exhibits strong order sensitivity, where reversing modality order leads to different reasoning outcomes despite identical multimodal content, as shown in Table~\ref{tab:topology_sensitivity}. 
We analyze the effect of modality ordering from the perspective of RoPE-based attention. Following the formulation of RoPE~\cite{su2024roformer}, consider an unrotated query representation $q_m$ at position $m$ and an unrotated key representation $k_n$ at position $n$. By the relative-position property of RoPE, their attention logit can be written as
\[
\ell(q_m,k_n)
=
q_m^T R_{n-m}k_n
=
\underbrace{q_m^T k_n}_{\text{content interaction}}
+
\underbrace{q_m^T(R_{n-m}-I)k_n}_{\text{positional modulation}},
\]
where $R_{n-m}=R_m^TR_n$ is the relative block-diagonal rotation matrix determined by the positional offset $n-m$. Thus, even when the underlying semantic content remains unchanged, the query-key interaction can be modulated by their relative positions.
Following the Abel transformation used in the original RoPE analysis~\cite{su2024roformer}, the attention interaction is bounded by
\[
\left|q_m^T R_{n-m}k_n\right|
\le
\sum_{i=0}^{d/2-1}
\left|S_{i+1}(n-m)\right|
\left|h_{i+1}-h_i\right|,
\]
where $h_i=q_m[2i:2i+1]k_n[2i:2i+1]$ and
$S_{i+1}(n-m)=\sum_{j=0}^{i}e^{\mathrm{i}(n-m)\theta_j}$, with the boundary convention $h_{d/2} := 0$.
We denote the right-hand side as $B(n-m)$. The RoPE analysis shows that $B(\cdot)$ exhibits a \textbf{long-term decaying} envelope as the relative distance increases. Hence,$
\left|\ell(q_m,k_n)\right|
\le B\left(|n-m|\right)$,
indicating that relative position imposes a distance-dependent upper envelope on the strength of query-key interaction.
For our question-to-modality analysis, let
$\Delta m_A=p_{Q,t}-p_{A,t}$ and
$\Delta m_V=p_{Q,t}-p_{V,t}$
denote the relative temporal positions between the question and the audio or visual modality, respectively. Under a modality-ordering intervention, the relative positions vary with the modality order. Therefore,
$A\rightarrow V\rightarrow Q:|\Delta m_V|<|\Delta m_A|$,
and $
V\rightarrow A\rightarrow Q:
|\Delta m_A|<|\Delta m_V|
$
. A smaller relative distance to the question corresponds to a less restrictive attenuation envelope, leaving the modality's query-key interaction less constrained and potentially allowing greater attention allocation. 

To verify this hypothesis, we compare semantic relevance and attention allocation under different modality organizations. 
Specifically, $S(Q,A)$ and $S(Q,V)$ measure the semantic similarity, and $A(Q,A)$ and $A(Q,V)$ denote the attention.
As shown in Table~\ref{tab:semantic_attention}, reversing modality order does not change semantic relevance. 
However, attention allocation changes substantially. 
In the $A\rightarrow V$ organization, visual tokens receive higher attention despite lower semantic relevance, while in the $V\rightarrow A$ organization, attention shifts toward audio tokens after they are placed closer to the question representation. 
Layer-wise attention visualization in Figure~\ref{fig:attention_change}(a) further shows that this positional preference consistently emerges across Transformer layers. We additionally visualize the corresponding attended regions before and after reversing modality order, and observe that the model's focus shifts (Figure~\ref{fig:attention_change}(b)).
These results suggest that modality accessibility is not solely determined by semantic importance, but also influenced by positional organization. 
Aligning the modality containing more relevant evidence with closer relative positions may enable more effective evidence utilization, leading to improved reasoning performance.

\begin{table*}[t!]
\centering
\footnotesize
\setlength{\tabcolsep}{3pt}
\vspace{-0.2cm}
\caption{\textbf{Semantic relevance vs. attention allocation between question and audio/visual on audio/video-bias subset on Qwen-Omni-7B.}}
\vspace{-0.3cm}
\label{tab:semantic_attention}
\begin{tabular*}{\linewidth}{@{\extracolsep{\fill}}ccccccc}
\toprule
\textbf{Subset}&\textbf{Order} 
& $S(Q,A)$ 
& $S(Q,V)$ 
& $A(Q,A)$ 
& $A(Q,V)$ &Acc
\\
\midrule
\multirow{2}{*}{Audio-Bias} & $A \to V$ &\textbf{0.234} &{0.151} &0.101 &\textbf{0.117} &48.0\\
& $V \to A$ &\textbf{0.234} &{0.151} &\textbf{0.152} & 0.071 &\textbf{48.8}\\\hline
\multirow{2}{*}{Video-Bias} & $A \to V$ &0.098 &{\textbf{0.114}} &0.079 &\textbf{0.129} &\textbf{50.0}\\
& $V \to A$ &0.098 &{\textbf{0.114}} &\textbf{0.126} & 0.089 &45.0\\
\bottomrule
\vspace{-0.6cm}
\end{tabular*}
\end{table*}

\begin{table*}[t]
\centering
\footnotesize
\setlength{\tabcolsep}{4pt}
\begin{minipage}{0.48\linewidth}
\centering
\caption{\textbf{Temporal reasoning analysis.}}
\vspace{-0.3cm}
\label{tab:temporal_reasoning}
\begin{tabular}{c|ccc}
\toprule
\textbf{Organization} &\textbf{Local Attn.} & \textbf{Global Attn.} &\textbf{Acc.} \\
\midrule
\textbf{Sequential} &0.00010  & 0.00019 & 45.4\\
\textbf{Interleaved} &\textbf{0.00033}  &0.00018 &\textbf{48.5}\\
\bottomrule
\end{tabular}
\end{minipage}
\hfill
\begin{minipage}{0.48\linewidth}
\centering
\caption{\textbf{Temporal hallucination analysis.}}
\vspace{-0.3cm}
\label{tab:temporal_ambiguity}
\begin{tabular}{c|ccc}
\toprule
\textbf{Organization} &\textbf{Local Attn.} & \textbf{Global Attn.} &\textbf{Acc.} \\
\midrule
\textbf{Sequential} &0.00021 &0.00020 &\textbf{75.0} \\
\textbf{Interleaved} &\textbf{0.00074} &0.00021 &67.9 \\
\bottomrule
\end{tabular}
\end{minipage}
\vspace{-0.6cm}
\end{table*}

\vspace{-0.3cm}
\paragraph{Temporal Reasoning}
For temporal reasoning, we observe that Interleaved organization achieves the best performance, as shown in Table~\ref{tab:topology_sensitivity}. 
Following the analysis above, the strength of cross-modal interaction between an audio chunk $A_i$ and its temporally corresponding visual chunk $V_i$ is subject to the distance-dependent upper bound $B\left(|n-m|\right)$. Therefore, the temporal organization directly determines the positional constraint imposed on their interaction.
Consider $T$ audio and visual chunks. Under the Sequential organization
$A_1\cdots A_TV_1\cdots V_T$, the corresponding chunks satisfy $\Delta p_i^{\mathrm{seq}}=p_{V_i}-p_{A_i}=T,$ whereas under the Interleaved organization $A_1V_1\cdots A_TV_T$, $\Delta p_i^{\mathrm{int}}=p_{V_i}-p_{A_i}=1$.
Thus,
$$
|\Delta p_i^{\mathrm{int}}|
<
|\Delta p_i^{\mathrm{seq}}|
\quad\Rightarrow\quad
|B_{\mathrm{int}}(1)|>|B_{\mathrm{seq}}(T)|.
$$
Hence, Interleaved organization may impose a less restrictive positional constraint on the temporally corresponding audio-visual chunks, allowing the model to allocate stronger attention to the local cross-modal evidence that is important for temporal reasoning.

To verify this hypothesis, we compare temporal attention patterns between Interleaved and Sequential organizations. 
Specifically, we measure the attention between temporally adjacent audio-video chunks. 
As shown in Table~\ref{tab:temporal_reasoning}, Interleaved organization consistently exhibits higher local temporal attention than Sequential organization, which aligns with its superior performance on temporal reasoning samples.
However, stronger local attention is not always beneficial. 
For temporally hallucination samples in Table~\ref{tab:temporal_ambiguity}, we observe that Interleaved still maintains higher local temporal attention but achieves lower performance than grouped organization, suggesting that excessive locality may introduce misleading temporal correlations.
This suggests that the locality bias introduced by RoPE acts as a task-dependent inductive bias: it benefits fine-grained temporal alignment but may cause over-reliance on local evidence and impair global temporal reasoning.

Please note that the Finding and experimental subsets serve different purposes and are constructed independently: the former characterizes topology effects under specific reasoning scenarios, and the latter evaluates CoM’s ability to identify effective modality organizations.

\vspace{-0.2cm}
\section{Chain-of-Modality: Topology-Controlled Dynamic Inference}
\begin{figure*}[!t]
    \centering
    \includegraphics[width=1\linewidth]{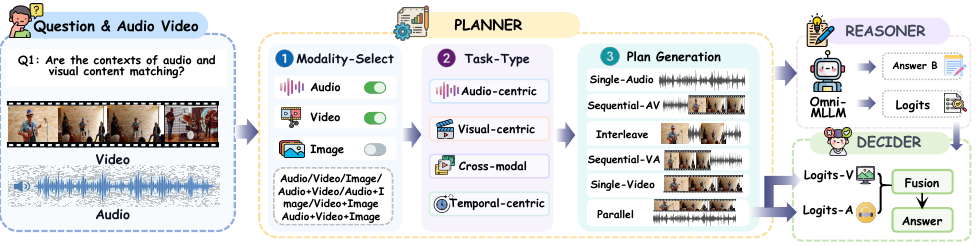}
    \vspace{-0.7cm}
   \caption{\textbf{Architecture of the CoM framework.} CoM is a topology-controlled inference framework, which consists of Planner, Reasoner, and Decider. }
    \label{fig:model}
    \vspace{-0.4cm}
\end{figure*}

\vspace{-0.2cm}
Our empirical analysis reveals that modality organization serves as an important factor in Omni-MLLM inference: different reasoning requirements favor different modality topologies, and no single organization consistently achieves optimal performance.
These findings motivate a natural question: can modality organization be dynamically controlled according to the underlying reasoning requirements of each input?
To answer this question, we introduce \textbf{Chain-of-Modality (CoM)} (Figure~\ref{fig:model}), a controllable inference framework that adaptively reorganizes modality interactions without modifying the underlying Omni-MLLM parameters.
CoM consists of three functional components: \textbf{Planner}, \textbf{Reasoner}, and \textbf{Decider}. 
Given a query with corresponding audio-visual inputs, the Planner selects an appropriate modality organization among \textit{Single}, \textit{Parallel}, \textit{Sequential}, and \textit{Interleaved}. 
The Reasoner performs topology-guided inference using the original Omni-MLLM backbone. 
For Parallel organization, where modality-specific predictions are independently obtained, the Decider aggregates option-level logits to produce the final prediction.
Formally, let $f_{\theta}$ denote the original Omni-MLLM backbone and $f_{\phi}$ denote the Planner. 
Different functional stages are instantiated through specialized system prompts, including $p_{plan}$ for topology selection and $p_{reason}$ for topology-guided reasoning.

\vspace{-0.2cm}
\subsection{Adaptive Modality Planning}
\vspace{-0.2cm}
The Planner explicitly models modality organization as an inference decision. Given a query $\bm{Q}$ and available modality set $\mathcal{M}=\{\bm{A},\bm{V},\bm{I}\}$, where $\bm{A}$, $\bm{V}$, and $\bm{I}$ denote audio, video, and image modalities, respectively, the Planner selects a task-dependent modality subset and organization strategy according to query requirements and available sensory evidence:
\begin{equation}
f_{\phi}(\bm{Q},\mathcal{M}, p_{plan})\rightarrow({t}_{think}, \mathcal{M}_{min},\mathcal{F}),
\end{equation}
where ${t}_{think}$ denotes the Planner reasoning output, $\mathcal{M}_{min}$ represents the selected modality subset, and $\mathcal{F}\in\{\text{Single},\text{Parallel},\text{Sequential},\text{Interleaved}\}$ denotes the selected modality organization strategy.
The selected organization determines how modality information is presented to the Reasoner. Specifically, interleaved organization constructs fine-grained cross-modal sequences, such as alternating modality representations $(\bm{A}_1,\bm{V}_1,\bm{A}_2,\bm{V}_2,\dots)$, and sequential organization arranges selected modalities according to explicit orders, such as $\bm{A}\rightarrow \bm{V}$ or $\bm{V}\rightarrow \bm{A}$. Parallel organization processes modalities independently and is suitable when modality-specific evidence can be aggregated without explicit cross-modal ordering. Single organization is selected when $\mathcal{M}_{min}$ contains only one modality, in which case the Reasoner performs unimodal inference directly following Equation~\ref{eq:reasoner}. By separating modality organization from multimodal reasoning, CoM enables explicit investigation of how different organization strategies influence inference behavior and whether effective modality organizations can be discovered through learning.

\vspace{-0.2cm}
\subsection{Modality-Organized Reasoning and Decision Fusion}
\vspace{-0.2cm}
After modality organization is determined by the Planner, the Reasoner performs inference according to the selected strategy. For sequential or interleaved organizations selected by $\mathcal{F}$, the Reasoner directly receives the organized modality input and generates the final prediction:
\begin{equation}
y = f_{\theta}(\mathcal{O}, \bm{Q}, p_{reason}),
\label{eq:reasoner}
\end{equation}
where $\mathcal{O}$ denotes the organized modality input constructed according to the Planner output.
For parallel organization, $\mathcal{M}_{min}$ is restricted to a pair of modalities, and CoM independently constructs two modality-specific branches
$y_i = f_{\theta}(\mathcal{O}_i, \bm{Q}, p_{reason}), \quad i=1,2,$
where each branch corresponds to an individual modality-specific input. When different branches produce consistent predictions, CoM directly adopts the shared prediction. Otherwise, the Decider resolves the disagreement by averaging the modality-specific prediction distributions:
\begin{equation}
p(y|\bm{Q},\mathcal{M})
=
\frac{1}{2}\sum_i p_i(y|\bm{Q},\mathcal{O}_i,p_{reason}),
\end{equation}
where $p_i(y|\bm{Q},\mathcal{O}_i,p_{reason})$ denotes the prediction distribution from the $i$-th modality branch. This design retains the Omni-MLLM backbone for reasoning while enabling explicit control over modality organization and aggregating heterogeneous modality evidence.
\begin{table*}[t!]
\centering
\footnotesize
\renewcommand{\tabcolsep}{5.5pt}
\vspace{-0.3cm}
\caption{\textbf{Comprehensive performance comparison on our defined subset and full benchmark.}  }
\label{tab:overall_comparison}
\begin{tabular}{ccccccccc}
\toprule
    \multirow{2}{*}{\textbf{Method}} & \multicolumn{3}{c}{\textbf{Music-AVQA}} & \multicolumn{2}{c}{\textbf{AVHBench}} & \multirow{2}{*}{\textbf{OmniBench}} & \multirow{2}{*}{\textbf{WorldSense}} &
    \multirow{2}{*}{\textbf{DailyOmni}} \\
 & A & V & A-V & A-Hal & V-Hal  & & &   \\
\midrule
\rowcolor{paleblue} \multicolumn{9}{c}{\textit{\textbf{Topology-sensitive Subset}}}\\
EchoInk &37.5 &29.4 &32.2 &49.2 &19.0 &25.4 &30.1 &34.1 \\
Omni-R1 &32.5 &19.6 &32.2 &30.9 &21.9 &30.6 &34.7 &37.6 \\
AVATAR &29.0 &18.2 &28.9 &30.9 &21.8 &31.0 &34.4 &36.5 \\
\midrule
\textbf{Qwen-Omni-7B} &31.5 &21.0 &28.5 &16.2 &25.9 &23.4 &27.8 &29.0 \\
\rowcolor{gray!5} \textbf{+CoM} &\textbf{35.8}  &\textbf{25.8} &\textbf{30.3} &\textbf{40.4} &\textbf{32.7} &\textbf{29.0} &\textbf{37.3} &\textbf{29.7} \\
\rowcolor{gray!5} \textbf{+CoM(SFT)} &\textbf{39.5} &\textbf{40.0}  &\textbf{39.7} &\textbf{51.7} &\textbf{49.2} &\textbf{35.9} &\textbf{42.6} &\textbf{38.3} \\\hline
\rowcolor{paleblue} \multicolumn{9}{c}{\textit{\textbf{Full Benchmark}}}\\
 \textbf{Qwen-Omni-7B} &72.6 &74.3 &57.8 &73.2 &67.1 &49.1 &\textbf{41.4} &55.6 \\
\rowcolor{paleblue} \rowcolor{gray!5} \textbf{+CoM(SFT)} &\textbf{72.8} &\textbf{74.6} &\textbf{58.3} &\textbf{76.5} &\textbf{73.3} &\textbf{50.3} &{41.3} &\textbf{57.0} \\
\bottomrule
\end{tabular}
\vspace{-0.2cm}
\label{tab:zero-shot}
\end{table*}

\vspace{-0.4cm}
\section{Experiment}
\vspace{-0.2cm}
\label{sec:experiment}
\paragraph{Evaluation Benchmarks.}  
We evaluate our framework on five representative benchmarks: \textbf{Music-AVQA} \cite{yang2022avqa}, \textbf{OmniBench} \cite{li2024omnibench}, \textbf{DailyOmni} \cite{zhou2025daily}, \textbf{WorldSense} \cite{hong2025worldsense}, and \textbf{AVHBench} \cite{sung2024avhbench}. 

\vspace{-0.2cm}
\paragraph{Implementation Details}
\label{sec:experiment}
Our evaluation follows a two-stage paradigm:
\textbf{(1)}
We first evaluate a training-free setting across three representative Omni-MLLMs.
\textbf{(2)}
To investigate whether effective modality organization strategies can be learned rather than manually specified, we further train a lightweight Planner based on Qwen-Omni-7B using supervised fine-tuning. The trajectories are automatically generated by Qwen-Omni-30B \cite{xu2025qwen3} from OmniVideo-100K \cite{cai2026omnivideo}, resulting in 5,791 modality planning samples. T
We fine-tune only the Planner with LoRA for one epoch using a learning rate of $5\times10^{-5}$. This setting evaluates whether modality organization policies can be learned from external supervision and generalized to unseen tasks and benchmarks. For all experiments, we set the decoding temperature to $T=0$ and fix the random seed to ensure deterministic evaluation and minimize stochastic variations.


\vspace{-0.4cm}
\subsection{Main Results}
\vspace{-0.2cm}
To systematically analyze how modality organization affects different reasoning behaviors, we construct topology-sensitive subsets: (1) samples where unimodal inputs achieve correct predictions while multimodal integration leads to degraded performance, and (2) samples where identical multimodal content produces different predictions under different modality organizations. \textit{These subsets are designed as a diagnostic evaluation to isolate topology-induced errors and quantify the extent to which our method can correct them. As our goal is not only to improve general Omni-MLLM accuracy, but also to establish that topology-induced failures are systematic, identifiable, and learnable.}
We first evaluate the training-free CoM strategy on these topology-sensitive subsets across five representative multimodal benchmarks. 
The vanilla baseline follows the A-V organization for consistency, as OmniBench \cite{li2024omnibench} consists of audio-image tasks.

\begin{figure*}[!t]
    \centering
    \includegraphics[width=1\linewidth]{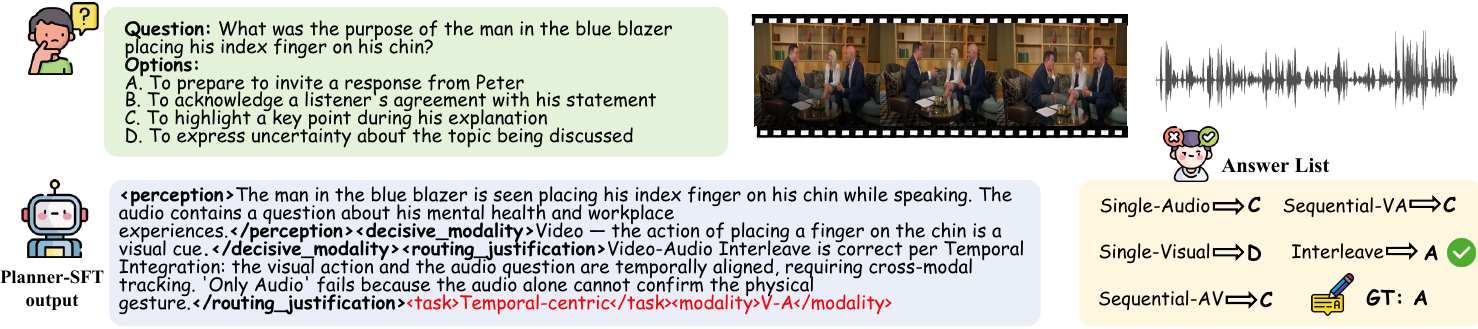}
    \vspace{-0.5cm}
  \caption{\textbf{Qualitative comparison of Planner-SFT and fixed modality organizations.} The results show that Planner-SFT can select task-adaptive modality sequences and improve multimodal reasoning.}
    \label{fig:quatitiva}
    \vspace{-0.5cm}
\end{figure*}

As shown in Table~\ref{tab:overall_comparison}, training-free CoM consistently improves over the vanilla Qwen-Omni-7B baseline across different benchmarks and reasoning scenarios. 
These results indicate that multimodal reasoning failures are not solely caused by insufficient perception capability, but can also arise from suboptimal organization of available evidence.
To investigate whether effective modality organization strategies can be learned, we further introduce a lightweight Planner-SFT stage. 
As shown in Table~\ref{tab:overall_comparison}, CoM(SFT) further improves over the training-free strategy on most benchmarks, demonstrating that modality organization policies can be learned through external supervision and transferred to downstream multimodal reasoning tasks.
We further compare CoM with existing omni-modal models built upon Qwen-Omni-7B \cite{kulkarni2025avatar, xing2025echoink, zhong2025omni}. We re-evaluate all models on the same topology-sensitive subsets using the same prompts and identical evaluation settings and CoM achieves competitive performance. Finally, we report results on the full benchmarks. CoM achieves comparable or better performance than the vanilla Qwen-Omni-7B, with gains comparable in magnitude to those reported by Omni-MLLMs trained with GRPO~\cite{kulkarni2025avatar}. CoM achieves these gains through inference-time modality organization without updating the underlying Omni-MLLM.
The lack of gain on WorldSense may stem from its relatively lower sensitivity to modality organization at the full-benchmark level. The gap between unimodal and multimodal organizations is only around 3 points, compared with nearly 20 points on DailyOmni and OmniBench, leaving less room for topology optimization.
Detailed qualitative examples of modality planning decisions and reasoning behaviors are provided in Figure~\ref{fig:quatitiva}.

\begin{table*}[t!]
\centering
\footnotesize

\begin{minipage}{0.45\linewidth}
\centering
\renewcommand{\tabcolsep}{0.3pt}
\vspace{-0.3cm}
\caption{\textbf{Generalization of CoM across different Omni-MLLM architectures and scales on topology-sensitive subset.}}
\label{tab:other_model}
\begin{tabular}{cccc|cccc}
\toprule
\multirow{1}{*}{\textbf{Method}} 
& \multicolumn{1}{c}{\textbf{Wor.}}
& \multicolumn{1}{c}{\textbf{Om.}}
& \multicolumn{1}{c|}{\textbf{Da.}} &\multirow{1}{*}{\textbf{Method}}  &\multicolumn{1}{c}{\textbf{Wor.}}
& \multicolumn{1}{c}{\textbf{Om.}}
& \multicolumn{1}{c}{\textbf{Da.}}\\
\midrule
\rowcolor{paleblue}
\rowcolor{paleblue} \multicolumn{8}{c}{\textit{\textbf{Topology-sensitive Subset}}}\\
Omni-3B & 28.3& 27.4&30.2 &Ola-7B &13.6 &15.2 &11.2 \\
+CoM &31.2 & 31.4&30.3 & +CoM &13.3 &17.8 &16.8\\
+Planner &\textbf{32.9} &\textbf{40.8} &\textbf{38.9} & +Planner &\textbf{28.5} &\textbf{32.2} &\textbf{22.2}\\\hline
\rowcolor{paleblue} \multicolumn{8}{c}{\textit{\textbf{Full Benckmark}}}\\
Omni-3B & 28.3& 27.4&30.2 &Ola-7B\\
+Planner &\textbf{32.9} &\textbf{40.8} &\textbf{38.9} &+Planner\\
\bottomrule
\end{tabular}
\vspace{-0.3cm}
\end{minipage}
\hfill
\begin{minipage}{0.48\linewidth}
\centering
\vspace{-0.3cm}
\label{fig:organization_distribution}
\includegraphics[width=\linewidth]{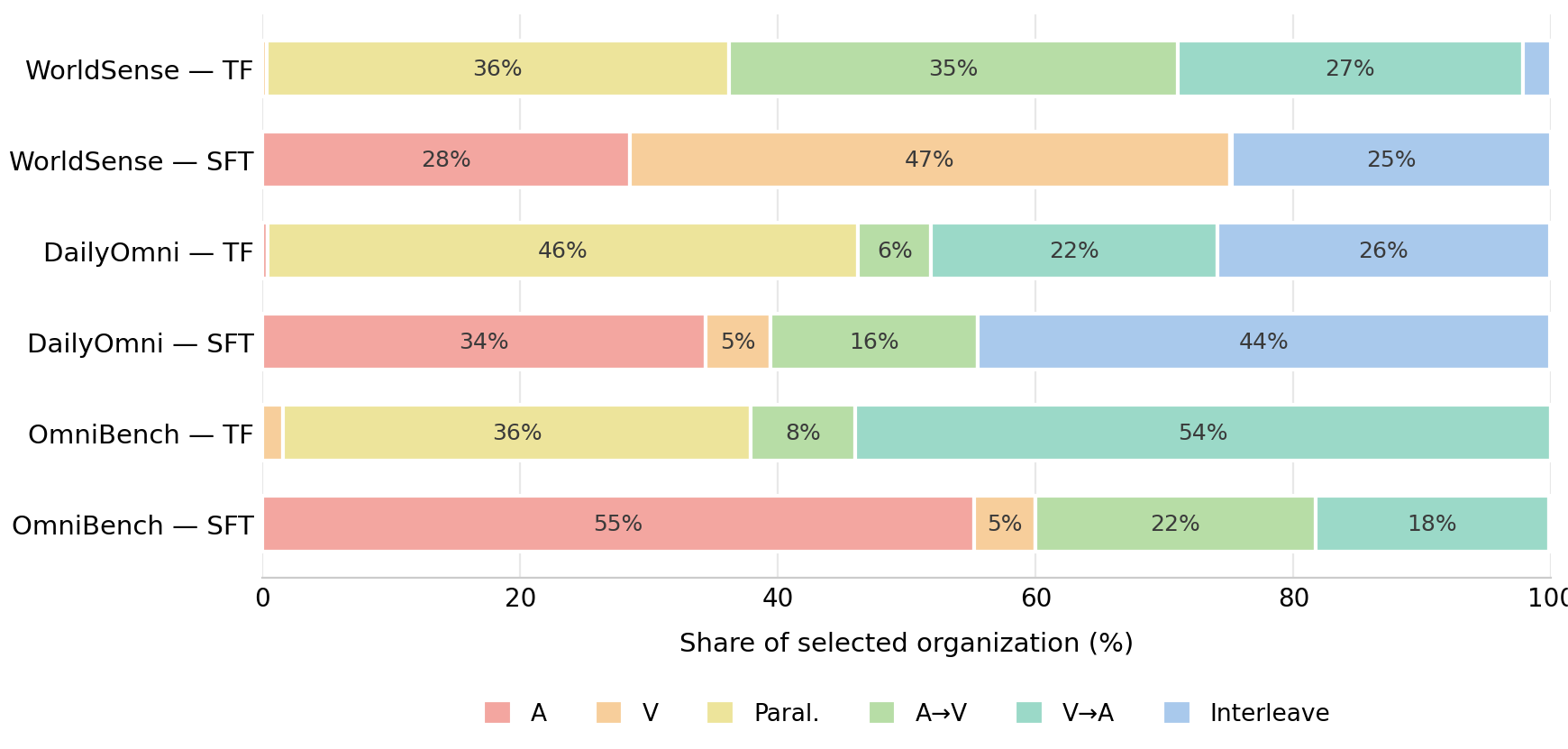}
\captionof{figure}{\textbf{Distribution of selected modality organizations across benchmarks under training-free (TF) and Planner-SFT (SFT) CoM.}\label{fig:organization_distribution}}
\vspace{-0.3cm}
\end{minipage}

\end{table*}


\vspace{-0.3cm}
\subsection{Cross-Model Generalization Analysis}
\vspace{-0.2cm}
To investigate whether the effectiveness of CoM depends on a specific model architecture or scale, we evaluate our framework on diverse Omni-MLLMs, including Qwen-Omni-3B \cite{xu2025qwen2} and Ola-7B \cite{liu2025ola}, which follow different Omni-MLLM architectures.
As shown in Table~\ref{tab:other_model}, training-free CoM improves the performance of both Qwen-Omni-3B and Ola-7B on the majority of evaluated benchmarks. These results demonstrate that the effectiveness of explicit modality organization is not restricted to a specific backbone architecture or model scale.
We further investigate whether the learned planning strategy can generalize across different Omni-MLLMs. Specifically, we transfer the Planner trained on Qwen-Omni-7B to other Omni-MLLMs while keeping their reasoning backbones unchanged. As shown in Table~\ref{tab:other_model}, the transferred Planner further improves performance across multiple models and benchmarks, suggesting that the learned modality organization policy generalizes beyond the source model rather than capturing model-specific behaviors (Appendix~\ref{sec:consistency}).

\vspace{-0.3cm}
\subsection{Adaptive Modality Organization Analysis}
\vspace{-0.2cm}
Beyond overall performance improvements, we further analyze how CoM selects modality organizations under different settings. 
Figure~\ref{fig:organization_distribution} summarizes the organization distributions selected by training-free CoM and Planner-SFT CoM. 
For the training-free setting, the selected organizations vary substantially across benchmarks, suggesting that modality organization preferences vary with sample characteristics.
After Planner-SFT, the organization distributions exhibit clearer sample-dependent preferences. Rather than maintaining a uniform routing strategy, the learned Planner shifts toward different organization patterns across samples.
These observations demonstrate that CoM performs adaptive modality orchestration rather than fixed topology selection. 

\vspace{-0.2cm}
\section{Conclusion}
\vspace{-0.2cm}
In this work, we reveal modality organization as an overlooked factor influencing Omni-MLLM inference. 
Our analysis shows that different reasoning scenarios favor different modality topologies, indicating that no single organization is universally optimal. We introduce Chain of Modality (CoM), which enables adaptive modality organization through inference-time planning. Experiments across benchmarks, architectures, and model scales establish that topology-induced failures are not merely stochastic artifacts, but exhibit systematic patterns that can be partially controlled.

\newpage

\bibliography{example_paper}
\bibliographystyle{iclr2027_conference}

\newpage

\end{document}